\documentclass[runningheads,a4paper]{llncs}

\usepackage{latexsym}
\usepackage{multirow}
\usepackage{amsfonts}
\usepackage{amssymb}
\usepackage{amsmath}
\usepackage{setspace}
\usepackage{enumitem}
\usepackage{graphicx}

\usepackage[normalem]{ulem}
\usepackage{verbatim}
\usepackage{enumitem}
\usepackage{wrapfig}

\setlist[itemize]{leftmargin=*}
\DeclareMathOperator{\softmax}{softmax}

\usepackage{url}
\urldef{\mailsa}\path|{alfred.hofmann, ursula.barth, ingrid.haas, frank.holzwarth,|
\urldef{\mailsb}\path|anna.kramer, leonie.kunz, christine.reiss, nicole.sator,|
\urldef{\mailsc}\path|erika.siebert-cole, peter.strasser, lncs}@springer.com|
\newcommand{\keywords}[1]{\par\addvspace\baselineskip
\noindent\keywordname\enspace\ignorespaces#1}

\begin{document}

\mainmatter  

\title{Patterns versus Characters\\in Subword-aware Neural Language Modeling}

\titlerunning{Patterns versus Characters}

%
%
\author{Rustem Takhanov \and Zhenisbek Assylbekov}
\authorrunning{Rustem Takhanov and Zhenisbek Assylbekov}

\institute{Nazarbayev University, Astana, Kazakhstan,\\
\email{\{rustem.takhanov, zhassylbekov\}@nu.edu.kz}}

%
%

\toctitle{Lecture Notes in Computer Science}
\tocauthor{Authors' Instructions}
\maketitle

\begin{abstract}
Words in some natural languages can have a composite structure. Elements of this structure include the root (that could also be composite), prefixes and suffixes with which various nuances and relations to other words can be expressed. Thus, in order to build a proper word representation one must take into account its internal structure. From a corpus of texts we extract a set of frequent subwords and from the latter set we select patterns, i.e. subwords which encapsulate information on character $n$-gram regularities. The selection is made using the pattern-based Conditional Random Field model~\cite{ye2009conditional,takhanov2013inference} with $l_1$ regularization.
Further, for every word we construct a new sequence over an alphabet of patterns.
The new alphabet's symbols confine a local statistical context stronger than the characters, therefore they allow better representations in ${\mathbb{R}}^n$ and are better building blocks for word representation. 
In the task of subword-aware language modeling, pattern-based models outperform character-based analogues by 2-20 perplexity points. Also, a recurrent neural network in which a word is represented as a sum of embeddings of its patterns is on par with a competitive and significantly more sophisticated character-based convolutional architecture.
\end{abstract}

\section{Introduction}
The goal of natural language modeling is, given a corpus of texts from a certain language, to build a probabilistic distribution over all possible sequences of words/sentences. Historically, first approaches to the problem \cite{Shannon,chomsky1956three} were highly interpretable, involving syntax and morphology, i.e. the internal structure of such models was of interest even to linguists. Nowadays the best performance is achieved by the so called recurrent neural network language models (RNNLM), which unfortunately lack the desired properties of interpretability. 

For rich-resource languages the amount of training data, i.e. a corpus of texts, is bounded only by the computational power of the language modeling method. Due to this, most of  RNNLM methods treat text as a sequence of token identifiers, where a token corresponds to either a word, or punctuation mark. Indeed, if any word appears in a text in various different contexts, a method can learn high quality word representation without taking into account its morphology. This logics fails when a corpus of texts is not large enough, and the problem is aggravated for morphology-rich languages, such as, e.g., turkic or finno-ugric languages. Thus, the problem of word representation that would take into account an internal structure of a word becomes very actual --- recent advances in language modeling are connected with treating words as sequences of characters or other subword units.

Much research has been done on character-level neural language modeling  \cite{mikolov2012subword,graves2013generating,ling-EtAl:2015:EMNLP2,kim2016character,lankinen2016character,verwimp2017character}. However, not much work exploits character $n$-grams that occur in a word. In \cite{sperr2013letter} a word is represented using a character $n$-gram count vector, followed by a single nonlinear transformation to yield a low-dimensional embedding; the word embeddings are then fed into neural machine translation models. In \cite{DBLP:conf/emnlp/WietingBGL16} a very similar technique is used and an evaluation on three other tasks (word similarity, sentence similarity, and part-of-speech tagging) is performed; they demonstrate that their method outperforms more complex architectures based on character-level recurrent and convolutional neural networks. 
Probably closest to ours is an approach from \cite{bojanowski2016enriching} where a word representation is a sum of terms, each term corresponding to a certain $n$-gram that occurs in that word. One weekness of the mentioned approaches is that all possible $n$-grams that occur in a corpus of texts are present there in an {\em a priori} equal way, and a difference in their value for word representation is calculated in the process of learning. Whereas we in advance select a subset of $n$-grams that could potentially enrich word vectors by subword information. For this purpose we use the pattern-based Conditional Random Field with $l_1$ regularization.

Our approach also differs in the following aspects: we (i) replace each character by a new symbol which in some way concentrates an information on previous characters, (ii) experiment with several ways of combining subword embeddings to produce word embeddings, and (iii) evaluate our methods on a ubiquitous language modeling task.

\section{A new alphabet for words}
Throughout the paper, we will use the following notation: if $\mathcal{X}$ is an alphabet, then $\mathcal{X}^\ast$ denotes a set of words over $\mathcal{X}$; for $\alpha, \beta\in \mathcal{X}^\ast$, $\alpha\beta$ denotes the concatenation of $\alpha$ and $\beta$; by $\ast$ we denote an arbitrary word.




\begin{figure}
\begin{center}
\includegraphics[scale=0.18]{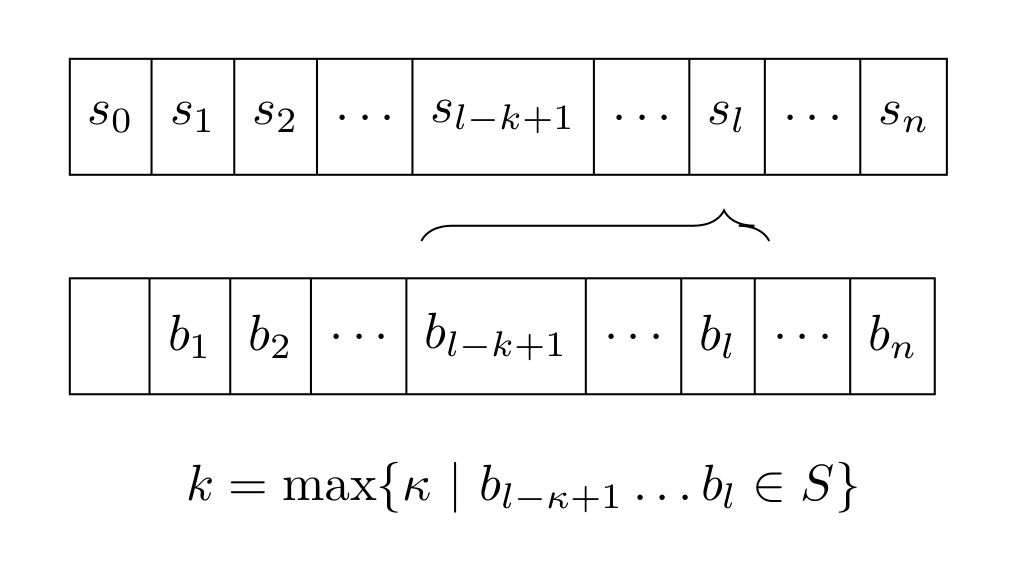}
\caption{Finite-State Machine.}
\label{fsm}
\end{center}
\end{figure}

The key trick that we use in this paper is replacing a word $a_1a_2\cdots a_k$ (that occurs in some context) over the initial alphabet $\mathcal{A}$ with a word $s_1 s_2
\cdots s_k$ over a new alphabet of states $\mathcal{S}$. Let us describe this substitution. We first define a finite state machine $\left(\mathcal{A}, \mathcal{S}, \delta, s_0\right)$, where $s_0$ is an initial state and $\delta:\mathcal{S}\times \mathcal{A}\rightarrow \mathcal{S}$ is a state-transition function. If we are given a sentence $\alpha = b_1 b_2 
\cdots b_{K}$ such that every $b_i$ is a character symbol from $\mathcal{A}$ (it could be a punctuation mark, i.e. a symbol that marks a boundary between words) our state machine reads this sentence and produces a sequence of states: $s_0 s_1 
\cdots s_K$. In the latter sequence, every $s_i$ corresponds to a state of our machine after reading a symbol $b_i$. Thus, every subsequence $b_i b_{i+1} \cdots b_{j}$ of the initial sentence $\alpha$ corresponds to a subsequence $s_i s_{i+1} \cdots s_{j}$ where $1\leq i \leq j \leq K$. Therefore, if $b_i b_{i+1} \cdots b_{j}$ corresponds to a word in a sentence $\alpha$, then we will substitute it with $s_i s_{i+1} \cdots s_{j}$.

Thus, given such a finite state machine, every word of a sentence can be rewritten over another alphabet $\mathcal{S}$. Let us describe now our finite state machine.

Suppose that after an analysis of a training set, i.e. of a corpus of texts from our language $\mathcal{L}$, we extract a certain finite set of sequences $\Pi_0\subseteq \mathcal{A}^\ast$ that we assume not only to be frequent, but in some way statistically characterising our language. A specific way of choosing $\Pi_0$ will be given in the following subsection. Any element $\pi\in \Pi_0$ we call {\em a pattern}. Any such set defines a set of states $\mathcal{S} = \{\beta| \mathop{\exists}\limits_{\pi\in\Pi_0}  \pi = \beta\ast\}$, which is, in fact, a set of all prefixes of patterns. We assume that an empty word $\varepsilon$ is also in $\mathcal{S}$ and define $s_0 = \varepsilon$.

Now we have to define a state-transition function $\delta$. Our idea is to construct it in such a way that after reading the first $l$ symbols of the sentence $b_1b_2 
\cdots b_l$ the machine should be in a state $s_l\in\mathcal{S}$ where $s_l$ is the longest word from $\mathcal{S}$ for which $b_1b_2 
\cdots b_l = \ast s_l$ (Figure \ref{fsm}). The latter decription  induces the following definition: for any $\alpha\in \mathcal{S}$ and $a\in \mathcal{A}$, $\delta(\alpha, a)$ is the longest word $\beta\in \mathcal{S}$ for which $\alpha a= \ast\beta$.

\subsection*{Patterns}
\begin{figure}
\begin{center}
\includegraphics[scale=0.18]{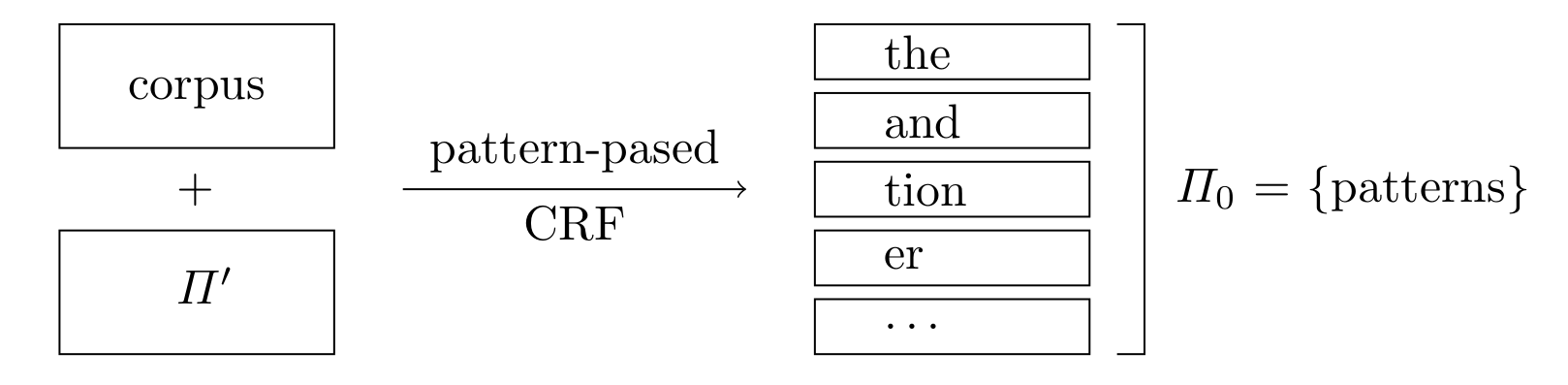}
\caption{Pattern mining.}
\label{pat_mining}
\end{center}
\end{figure}

In this subsection we will describe how we extract a set of patterns $\Pi_0$ from a corpus of texts (Figure \ref{pat_mining}). By a corpus of texts we understand a training set $T=\{\alpha_1, \cdots, \alpha_L\}\subseteq \mathcal{A}^\ast$ where $\alpha_i$ is a sentence from our language $\mathcal{L}$.


First we extract from our training set $T$ a set of patterns $\Pi'$ based on the following simple procedure: we fix in advance a threshold $f$ and put to $T$ only those words $\alpha\in \mathcal{A}^\ast$ that occur in $T$ in more than $f$ places. Then we apply a reduction procedure, i.e. if a) $\alpha$ is a subword of $\beta$, b) $\alpha$ and $\beta$ always occur together in $T$, then we delete $\alpha$ from $\Pi'$.
A pattern-based conditional random field model for our language is the following probability distribution over $\mathcal{A}^\ast$ \cite{ye2009conditional,takhanov2013inference}:

$$\Pr(b_1 \cdots b_K) = A\cdot e^{-E(b_1 \cdots b_K)},$$
where 
$E(b_1 \cdots b_K) = \sum_{\alpha\in\Pi'}\,\,\,\sum_{i<j: b_i
\cdots b_j = \alpha}c^\alpha,$
and $c^{\alpha}$, $\alpha\in\Pi'$, are parameters to be learned from $T$.

The learning is done by the minimization of the negative log-likelihood with $L_1$-regularization:
\begin{equation}\textstyle
-\sum_{i=1}^L\log \Pr(\alpha_i) + C\sum_{\alpha\in \Pi'}|c^\alpha|.\label{pat_learn}
\end{equation}
The latter function is convex, an efficient computation of its value and gradient is described in \cite{takhanov2013inference}. For the optimization we used the Limited-memory Broyden-Fletcher-Goldfarb-Shanno (L-BFGS) method written by Jorge Nocedal.
Via the parameter $C$ one can manage the number of patterns $\alpha\in \Pi'$ for which $c^\alpha\ne 0$. Finally, we define $\Pi_0 = \{\alpha\in \Pi' | c^\alpha\ne 0\}$.

\section{Subword-aware neural language model}\label{model}
\begin{figure}
\begin{center}
\includegraphics[scale=0.2]{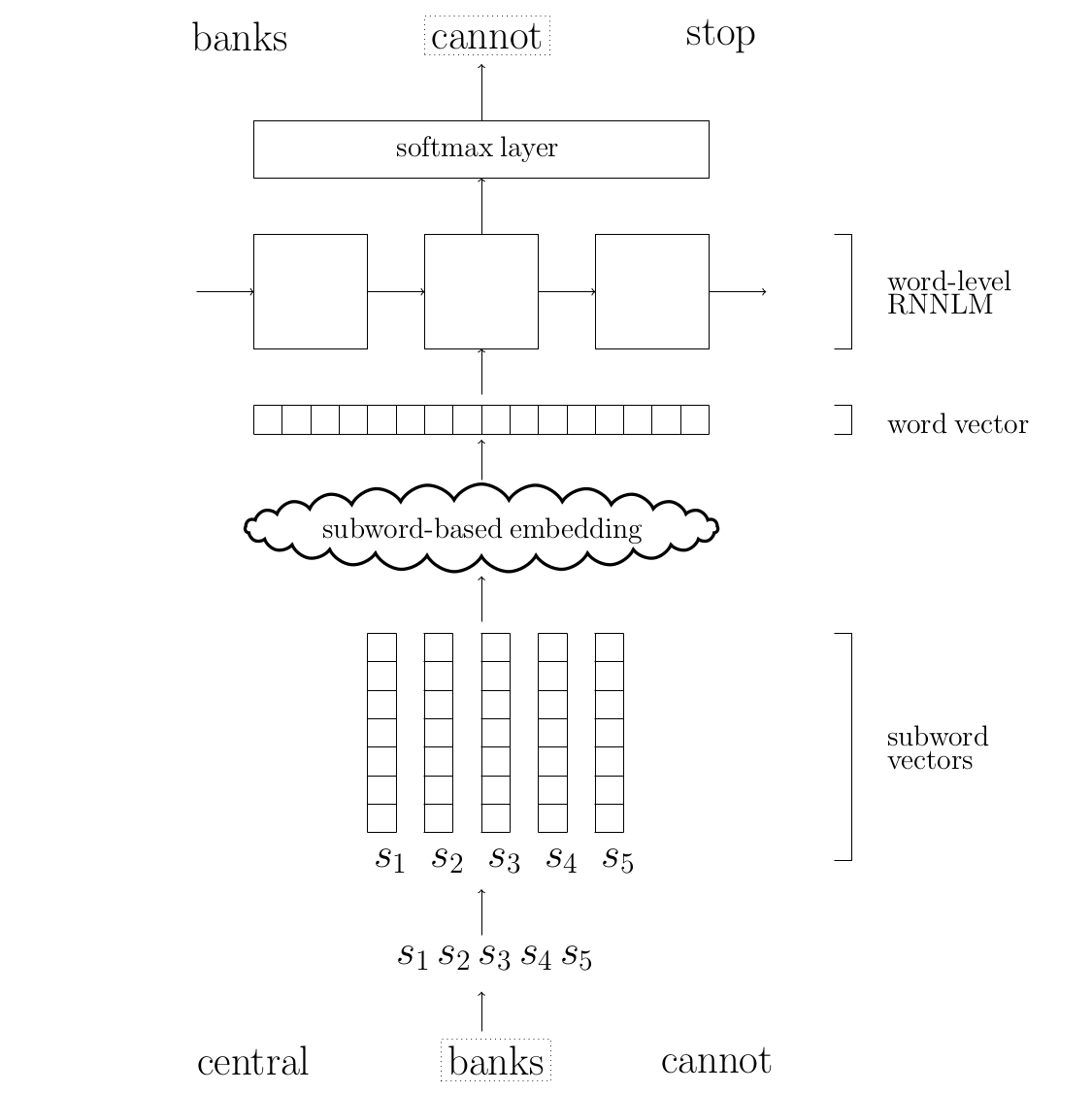}
  \caption{Subword-aware language model.}
  \label{architecture}
\end{center}
\end{figure}

In what follows, both regular characters and patterns are referred to as \textit{subwords}. The overall architecture of the subword-aware neural language model is displayed in Figure \ref{architecture}.

It consists of three main parts: (i) subword-based word embedding model, (ii) word-level recurrent neural network language model (RNNLM), and (iii) softmax layer. Below we describe each part in more detail. 

\noindent\textbf{Subword-based word embeddings:} A word $w\in\mathcal{W}$ (in a sentence) is defined by the sequence of its subwords $s_1\dots s_{n_w}\in\mathcal{X}^\ast$ ($\mathcal{X}=\mathcal{A}$ in the case of character-based representation, and $\mathcal{X}=\mathcal{S}$ in our pattern-based approach), and  each state is embedded into $d_\mathcal{X}$-dimensional space via an embedding matrix $\mathbf{E}_\mathcal{X}^{\text{in}}\in\mathbb{R}^{|X|\times d_\mathcal{X}}$ to obtain a sequence of state vectors:
\begin{equation}
\mathbf{s_1},\ldots,\mathbf{s_{n_w}}.\label{subw_vectors}
\end{equation}
Then we try three different methods to get an embedding of the word $w$:
\begin{itemize}
\item \textbf{Concat:} A simple concatenation of state vectors (\ref{subw_vectors}) into a single word vector:
$$\textstyle
\mathbf{w} = [\mathbf{s_1}; \mathbf{s_2}; \ldots; \mathbf{s_{n_w}}; \underbrace{\mathbf{0}; \mathbf{0}; \ldots; \mathbf{0}}_{n-n_w}].
$$
We either truncate (if $w$ consists of more than $n$ symbols) or zero-pad $\mathbf{w}$ so that all word vectors have the same length $n\cdot d_\mathcal{X}$ to allow batch processing. This approach is motivated by a desire to keep all the information regarding subwords, including the order in which they appear in the word.
\item \textbf{Sum:} A summation of subword vectors:
\begin{equation}\textstyle
\mathbf{w} = \sum_{t=1}^{n_w}\mathbf{s_t}.\label{sum}
\end{equation}
This approach was used by \cite{botha2014compositional} to combine a word and its morpheme embeddings into a single word vector.

\item \textbf{CNN:} A convolutional model of \cite{kim2016character}:
$$
\mathbf{w} = \text{CNN}(\mathbf{s_1},\ldots,\mathbf{s_{n_w}}).
$$
This method has already demonstrated excellent performance for character-level inputs, therefore we decided to apply it to patterns as well.
\end{itemize}

To model interactions between subwords, we feed the resulting word embedding $\textbf{w}$ into a stack of two highway layers \cite{srivastava2015training} with dimensionality $d_\text{HW}$  per layer. In cases when dimensionality of $\textbf{w}$ does not match $d_\text{HW}$, we project it into $\mathbb{R}^{d_\text{HW}}$.

\noindent\textbf{Word-level RNNLM}: Once we have embeddings $\mathbf{w_{1:k}}$ for a sequence of words $w_{1:k}$, we can use a word-level RNN language model to produce a sequence of states $\mathbf{h_{1:k}}\in\mathbb{R}^{d_\text{LM}}$ according to
\begin{equation}
\mathbf{h_{t}} = \text{RNNCell}(\mathbf{w_t},\notag \mathbf{h_{t-1}}),\qquad 
\mathbf{h_0}=\mathbf{0}.\label{RNN}
\end{equation}
There is a big variety of RNN cells to choose from. The most advanced recurrent neural architectures, at the time of this writing, are RHN \cite{zilly2016recurrent} and NAS \cite{zoph2016neural}. However, to make our results directly comparable to the previous work of \cite{kim2016character} on character-level language modeling  we select a more conventional architecture -- a stack of two LSTM cells \cite{hochreiter1997long}.

\noindent\textbf{Softmax:} The last state $\mathbf{h_k}$ from (\ref{RNN}) is further used to predict the next word $w_{k+1}$ according to the probability distribution
\begin{equation}
\Pr(w_{k+1}|w_{1:k})=\softmax(\mathbf{h_k}\mathbf{W} + \mathbf{b}),\label{softmax}
\end{equation}
where $\mathbf{W}\in\mathbb{R}^{d_{\text{LM}}\times|\mathcal{W}|}$, $\mathbf{b}\in\mathbb{R}^{|\mathcal{W}|}$, and $d_{\text{LM}}$ is a hidden layer size of the RNN.

\section{Experimental Setup}\label{exp_setup}
\noindent\textbf{Data sets:} All models are trained and evaluated on the English PTB data set \cite{marcus1993building} utilizing the
standard training (0-20), validation (21-22), and test (23-24)
splits along with pre-processing by \cite{mikolov2010recurrent}. Since the PTB is criticized for being small nowadays, we also provide an evaluation on the WikiText-2 data set \cite{merity2016pointer}, which is approximately two times larger than PTB in size and three times larger in vocabulary. We do not append any additional symbols at the end of each line in WikiText-2, but remove spaces between equality signs in the sequences ``= ='' and ``= = ='', which occur in section titles.

\noindent\textbf{Hyperparameters:} The regularization parameter $C$ from (\ref{pat_learn}) is set to 1600, which results in 883 unique patterns ($|\Pi_0|=883, |\mathcal{S}|=890$) for the PTB data set (cf. 48 plain characters) and 1440 unique patterns ($|\Pi_0|=1440, |\mathcal{S}|=1471$) for the WikiText-2 data set (cf. 281 plain characters). We set the threshold value $f$ to 300 on the PTB and to 700 on the WikiText-2. We experiment with two configurations for the state size $d_\text{LM}$ of the word-level RNNLM: 300 (small models) and 650 (medium-sized models). Specification of other hyperparameters is given below. 

{\em Concat:} $d_\mathcal{A}=15$ (for characters), and $d_\mathcal{S}=30$ (for patterns). We give higher dimensionality to patterns as their amount significantly exceeds the amount of characters. $n$ is set to the 95$^\text{th}$ percentile of word lengths, i.e. 95\% of all words  have not more than $n$ characters\footnote{word length in characters and in patterns is the same.}. We do not set $n=\max_{w\in\mathcal{W}}{n_w}$, as this would result in excessive zero-padding. $d_\text{HW}=d_\text{LM}$.

{\em Sum:} $d_\mathcal{X}=d_\text{HW}=d_\text{LM}\in\{300, 650\}$ for both characters and patterns. We give higher dimensionality to subword vectors here (compared to other models) since the resulting word vector will have the same size as subword vectors (see (\ref{sum})).

{\em CNN:} In character-based models we choose the same values for hyperparameters as in the work of \cite{kim2016character}. For pattern-based models we choose: $d_\mathcal{S}=50$ and $d_\mathcal{S}=100$ for small and medium-sized models; filter widths are [1, 2, 3, 4, 5, 6] and [1, 2, 3, 4, 5, 6, 7] for small and medium-sized models; the corresponding depths (number of features per width) are [100, 50, 75, 100, 100, 100] and [100, 100, 150, 200, 200, 200, 200]. $d_\text{HW}=\sum\text{depths}\in\{525,1150\}$.

\noindent\textbf{Optimization} is done similarly to  \cite{zaremba2014recurrent,kim2016character,gal2016theoretically}. 
Training the models involves minimizing the negative log-likelihood over the corpus $w_{1:K}$: 
$$\textstyle
-\sum_{k=1}^K\log\Pr(w_k|w_{1:k-1})\longrightarrow\min,
$$
which is typically done by truncated BPTT \cite{werbos1990backpropagation,graves2013generating}. We backpropagate for 35 time steps using stochastic gradient descent where the learning rate is initially set to 0.7 and halved if the perplexity does not decrease on the validation set after an epoch. We use a batch size of 20. We train for 65 epochs, picking the best performing model on the validation set. Parameters of the models are randomly initialized uniformly in $[-0.05, 0.05]$, except the forget bias of the word-level LSTM, which is initialized to $1$, and the transform bias of the highway, which is initialized to values near $-2$. For regularization we use variational dropout \cite{gal2016theoretically} with dropout rates for small/medium Concat, Sum/medium CNN models as follows: 0.1/0.15/0.2 for the embedding layer, 0.2/0.3/0.35 for the input to the gates, 0.1/0.15/0.2 for the hidden units, and 0.2/0.3/0.35 for the output activations. We clip the norm of the gradients (normalized by minibatch size) at 5.

\section{Results}
The results of evaluation on PTB and WikiText-2 are reported in Tables \ref{results_ptb} and \ref{results_wikitext2} correspondingly. As one can see, models which process patterns consistently outperform those which use characters under small parameter budgets. However, the difference in performance is less pronounced when we allow more parameters.

\begingroup
\setlength{\tabcolsep}{3pt}
\begin{table}[h]
\caption{Results on the PTB for small (left) and medium-sized models.} 
\begin{center}
\begin{tabular}{| l | c c | c c |}
\hline
\parbox[t]{0.5mm}{\multirow{2}{*}{Model}} &  \multicolumn{2}{c|}{Characters} & \multicolumn{2}{c|}{Patterns} \\
      & Size & PPL & Size & PPL \\
\hline
Concat & 5M & 119.2 & 5M & \textbf{99.6}\\
Sum & 5M & 108.2 & 5M & \textbf{87.4} \\
CNN & 6M & 87.3 & 6M & \textbf{84.8} \\
\hline
\end{tabular}
\quad
\begin{tabular}{| l | c c | c c |}
\hline
\parbox[t]{0.5mm}{\multirow{2}{*}{Model}} &  \multicolumn{2}{c|}{Characters} & \multicolumn{2}{c|}{Patterns} \\
      & Size & PPL & Size & PPL \\
\hline
Concat & 15M & 91.5 & 15.8M & \textbf{83.6}\\
Sum & 15M & 91.5 & 15.5M & \textbf{82.1} \\
CNN & 20M & 79.6 & 20.5M & \textbf{77.2} \\
\hline
\end{tabular}
\end{center}
\label{results_ptb}
\end{table}
\begin{table}[h]
\caption{Results on WikiText-2 for small (left) and medium-sized models.} 
\begin{center}
\begin{tabular}{| l | c c | c c |}
\hline
\parbox[t]{0.5mm}{\multirow{2}{*}{Model}} &  \multicolumn{2}{c|}{Characters} & \multicolumn{2}{c|}{Patterns} \\
      & Size & PPL & Size & PPL \\
\hline
Concat & 11.9M & 138.2 & 12.1M & \textbf{114.2}\\
Sum & 11.9M & 124.0 & 12.3M & \textbf{101.9} \\
CNN & 12.9M & 105.2 & 13.0M & \textbf{102.8} \\
\hline
\end{tabular}
\quad
\begin{tabular}{| l | c c | c c |}
\hline
\parbox[t]{0.5mm}{\multirow{2}{*}{Model}} &  \multicolumn{2}{c|}{Characters} & \multicolumn{2}{c|}{Patterns} \\
      & Size & PPL & Size & PPL \\
\hline
Concat & 30.2M & 115.9 & 30.8M & \textbf{99.0}\\
Sum & 30.3M & 106.7 & 31.1M & \textbf{94.9} \\
CNN & 34.5M & 97.38 & 35.7M & \textbf{94.2} \\
\hline
\end{tabular}
\label{results_wikitext2}
\end{center}
\end{table}
\endgroup

Also, it is clearly seen that patterns are more beneficial for simple models, such as Concat and Sum, but have less effect on the CNN model, which shrinks the gap between characters and patterns. This is quite natural as patterns carry some information on character $n$-grams and, hence, can be considered as ``discrete convolutions'', which makes CNN over patterns not as efficient as CNN over regular characters. However, we notice that in all cases a simple sum of pattern embeddings (Pat-Sum) is on par with a more sophisticated convolution over character embeddings (Char-CNN). Faster\footnote{around 1.2x speedup on NVIDIA Titan X (Pascal)} training of the Pat-Sum  compared to the Char-CNN makes the patterns even more advantageous.

\noindent\textbf{Why does Pat-Sum perform equally well as Char-CNN?} As was described in Section \ref{model} word embeddings are processed by the two highway layers before they are fed into the RNNLM. Highway is a weighted average between nonlinear and identity transformations of the incoming word embedding:
$$
\mathbf{w} \mapsto \mathbf{t}\odot\sigma(\mathbf{w}\mathbf{A}+\mathbf{b})+(\mathbf{1}-\mathbf{t})\odot\mathbf{w},
$$ 
where $\mathbf{t}$, $\mathbf{A}$ and $\mathbf{b}$ are trainable parameters, $\sigma(\cdot)$ is a non-linear activation, $\mathbf{1}$ is a vector whose all components are 1 and $\odot$ is an operation of component-wise multiplication. The ideal input for the highway is the one that does not need to undergo a nonlinear transformation, i.e the highway will then be close to an identity operator, and hence in the ideal case we shall have $\mathbf{t}=\mathbf{0}$. But if $\mathbf{w}$ is rather ``raw'', then the highway should prepare it for the RNN (resulting in $\mathbf{t}\ne\mathbf{0}$). Such extra nonlinearity can measured by the closeness of $\mathbf{t}$ to $\mathbf{1}$. We hypothesize that the reason why
Pat-Sum performs well is that the sum of  pattern embeddings is \textit{already} a good word representation. Hence the highway in Pat-Sum does less nonlinear work than in Char-CNN: In Pat-Sum it is almost an identical transformation, and such a simple highway is well-trained according to~\cite{hardt2016identity}. 
\begin{figure}
\includegraphics[scale=0.33]{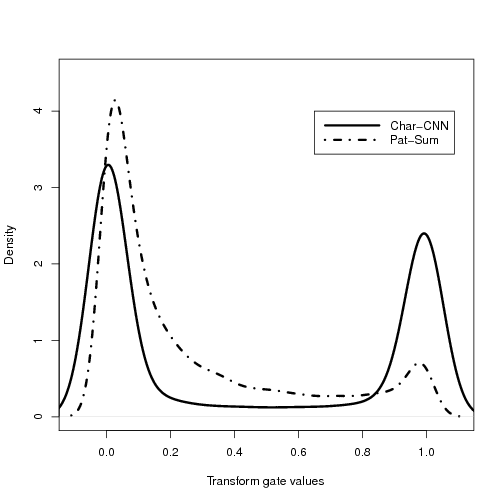}
\ \ 
\includegraphics[scale=0.33]{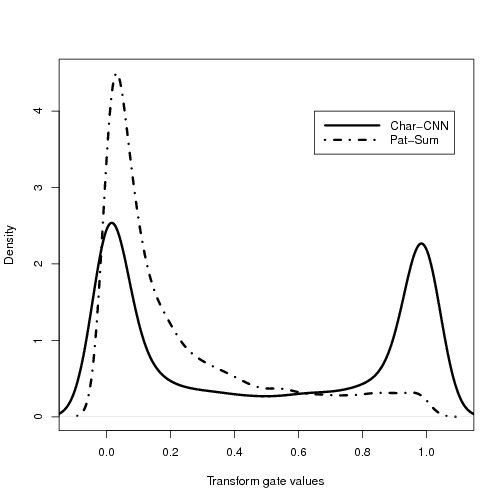}
\caption{Kernel density estimations of the transform gate values of the first (left) and second highway layers in Char-CNN and Pat-Sum.}
\label{transform_gates}
\end{figure}
To validate our hypothesis we compare the distributions of the transform gate $\mathbf{t}$ values from both highway layers of Pat-Sum and Char-CNN. The density plots in Fig. \ref{transform_gates} support our hypothesis: Pat-Sum does not utilize much of nonlinearity in the highway layers, while Char-CNN heavily relies on it.

\noindent\textbf{Source code:} All models were implemented in TensorFlow \cite{abadi2016tensorflow} and the source code for Pat-Sum is available at \url{https://github.com/zh3nis/pat-sum}.

\section{Conclusion}
Regular characters are rather uninformative when their embeddings are concatenated or summed to produce word vectors, but patterns, on the contrary, carry enough information to make these methods work significantly better. Convolutions over subword embeddings do capture $n$-gram regularities and, therefore, make the difference between characters and patterns less noticeable. It is noteworthy, that a simple and fast sum of pattern embeddings is on par with more sophisticated and slower convolutions over characters embeddings.

\section{Acknowledgments}
We gratefully acknowledge the support of NVIDIA Corporation with the donation of the Titan X Pascal GPU used for this research. 

\bibliography{pat-cnn}

\begin{thebibliography}{10}

\bibitem{abadi2016tensorflow}
Abadi, M., Agarwal, A., Barham, P., Brevdo, E., Chen, Z., Citro, C., Corrado,
  G.S., Davis, A., Dean, J., Devin, M.,  et~al.:
\newblock Tensorflow: Large-scale machine learning on heterogeneous distributed
  systems.
\newblock arXiv preprint arXiv:1603.04467 (2016)

\bibitem{bojanowski2016enriching}
Bojanowski, P., Grave, E., Joulin, A., Mikolov, T.:
\newblock Enriching word vectors with subword information.
\newblock arXiv preprint arXiv:1607.04606 (2016)

\bibitem{botha2014compositional}
Botha, J., Blunsom, P.:
\newblock Compositional morphology for word representations and language
  modelling.
\newblock In: Proceedings of the 31st International Conference on Machine
  Learning (ICML-14). (2014)  1899--1907

\bibitem{chomsky1956three}
Chomsky, N.:
\newblock Three models for the description of language.
\newblock IRE Transactions on information theory \textbf{2}(3) (1956)  113--124

\bibitem{gal2016theoretically}
Gal, Y., Ghahramani, Z.:
\newblock A theoretically grounded application of dropout in recurrent neural
  networks.
\newblock In: Advances in Neural Information Processing Systems. (2016)
  1019--1027

\bibitem{graves2013generating}
Graves, A.:
\newblock Generating sequences with recurrent neural networks.
\newblock arXiv preprint arXiv:1308.0850 (2013)

\bibitem{hardt2016identity}
Hardt, M., Ma, T.:
\newblock Identity matters in deep learning.
\newblock arXiv preprint arXiv:1611.04231 (2016)

\bibitem{hochreiter1997long}
Hochreiter, S., Schmidhuber, J.:
\newblock Long short-term memory.
\newblock Neural computation \textbf{9}(8) (1997)  1735--1780

\bibitem{kim2016character}
Kim, Y., Jernite, Y., Sontag, D., Rush, A.M.:
\newblock Character-aware neural language models.
\newblock In: Proceedings of the Thirtieth AAAI Conference on Artificial
  Intelligence, AAAI Press (2016)  2741--2749

\bibitem{lankinen2016character}
Lankinen, M., Heikinheimo, H., Takala, P., Raiko, T., Karhunen, J.:
\newblock A character-word compositional neural language model for finnish.
\newblock arXiv preprint arXiv:1612.03266 (2016)

\bibitem{ling-EtAl:2015:EMNLP2}
Ling, W., Dyer, C., Black, A.W., Trancoso, I., Fermandez, R., Amir, S., Marujo,
  L., Luis, T.:
\newblock Finding function in form: Compositional character models for open
  vocabulary word representation.
\newblock In: Proceedings of the 2015 Conference on Empirical Methods in
  Natural Language Processing, Lisbon, Portugal, Association for Computational
  Linguistics (September 2015)  1520--1530

\bibitem{marcus1993building}
Marcus, M.P., Marcinkiewicz, M.A., Santorini, B.:
\newblock Building a large annotated corpus of english: The penn treebank.
\newblock Computational linguistics \textbf{19}(2) (1993)  313--330

\bibitem{merity2016pointer}
Merity, S., Xiong, C., Bradbury, J., Socher, R.:
\newblock Pointer sentinel mixture models.
\newblock In: Proceedings of ICLR 2017. (2017)

\bibitem{mikolov2010recurrent}
Mikolov, T., Karafi{\'a}t, M., Burget, L., Cernock{\`y}, J., Khudanpur, S.:
\newblock Recurrent neural network based language model.
\newblock In: Interspeech. Volume~2. (2010) ~3

\bibitem{mikolov2012subword}
Mikolov, T., Sutskever, I., Deoras, A., Le, H.S., Kombrink, S., Cernocky, J.:
\newblock Subword language modeling with neural networks.
\newblock preprint (http://www. fit. vutbr. cz/imikolov/rnnlm/char. pdf) (2012)

\bibitem{Shannon}
Shannon, C.E., Weaver, W.:
\newblock A mathematical theory of communication.
\newblock (1963)

\bibitem{sperr2013letter}
Sperr, H., Niehues, J., Waibel, A.:
\newblock Letter n-gram-based input encoding for continuous space language
  models.
\newblock In: Proceedings of the Workshop on Continuous Vector Space Models and
  their Compositionality. (2013)  30--39

\bibitem{srivastava2015training}
Srivastava, R.K., Greff, K., Schmidhuber, J.:
\newblock Training very deep networks.
\newblock In: Advances in neural information processing systems. (2015)
  2377--2385

\bibitem{takhanov2013inference}
Takhanov, R., Kolmogorov, V.:
\newblock Inference algorithms for pattern-based crfs on sequence data.
\newblock In: ICML (3). (2013)  145--153

\bibitem{verwimp2017character}
Verwimp, L., Pelemans, J., Wambacq, P.,  et~al.:
\newblock Character-word lstm language models.
\newblock In: Proceedings of EACL 2017. (2017)

\bibitem{werbos1990backpropagation}
Werbos, P.J.:
\newblock Backpropagation through time: what it does and how to do it.
\newblock Proceedings of the IEEE \textbf{78}(10) (1990)  1550--1560

\bibitem{DBLP:conf/emnlp/WietingBGL16}
Wieting, J., Bansal, M., Gimpel, K., Livescu, K.:
\newblock Charagram: Embedding words and sentences via character n-grams.
\newblock In: Proceedings of the 2016 Conference on Empirical Methods in
  Natural Language Processing, {EMNLP} 2016, Austin, Texas, USA, November 1-4,
  2016. (2016)  1504--1515

\bibitem{ye2009conditional}
Ye, N., Lee, W.S., Chieu, H.L., Wu, D.:
\newblock Conditional random fields with high-order features for sequence
  labeling.
\newblock In: Advances in Neural Information Processing Systems. (2009)
  2196--2204

\bibitem{zaremba2014recurrent}
Zaremba, W., Sutskever, I., Vinyals, O.:
\newblock Recurrent neural network regularization.
\newblock arXiv preprint arXiv:1409.2329 (2014)

\bibitem{zilly2016recurrent}
Zilly, J.G., Srivastava, R.K., Koutn{\'\i}k, J., Schmidhuber, J.:
\newblock Recurrent highway networks.
\newblock arXiv preprint arXiv:1607.03474 (2016)

\bibitem{zoph2016neural}
Zoph, B., Le, Q.V.:
\newblock Neural architecture search with reinforcement learning.
\newblock In: Proceedings of ICLR 2017. (2017)

\end{thebibliography}
\bibliographystyle{splncs_srt}

\end{document}